\documentclass[runningheads]{llncs}

\usepackage{eccv}

\usepackage{eccvabbrv}

\usepackage{graphicx}
\usepackage{booktabs}

\usepackage[accsupp]{axessibility}  

\usepackage{amssymb}
\usepackage{pifont}
\usepackage{multirow} 
\usepackage{makecell}
\usepackage{gensymb}
\usepackage{footnote}
\usepackage{pgfplots}
\usepackage{caption}

\usepackage[pagebackref,breaklinks,colorlinks,citecolor=eccvblue]{hyperref}

\usepackage{orcidlink}

\begin{document}

\title{TexDreamer: Towards Zero-Shot High-Fidelity 3D Human Texture Generation} 

\titlerunning{TexDreamer}


\author {Yufei Liu\inst{1 \footnotemark[4] \footnotemark[5]},
Junwei Zhu\inst{2 \footnotemark[5]},
Junshu Tang\inst{3},
Shijie Zhang\inst{4},
Jiangning Zhang\inst{2},\\
Weijian Cao\inst{2},
Chengjie Wang\inst{2},
Yunsheng Wu\inst{2},
Dongjin Huang\inst{1 \footnotemark[6]}}

\institute{Shanghai University, Shanghai, China \and
Tencent Youtu Laboratory \and
Shanghai Jiao Tong University, Shanghai, China \and
Fudan University, Shanghai, China \\
\url{https://ggxxii.github.io/texdreamer/}
}

\authorrunning{Y. Liu~et al.}


\maketitle

\renewcommand{\thefootnote}{\fnsymbol{footnote}}
\footnotetext[4]{Work is done during the internship at Tencent YouTu Lab.} 
\footnotetext[5]{Co-first author.} 
\footnotetext[6]{Corresponding author.}

\begin{abstract}


Texturing 3D humans with semantic UV maps remains a challenge due to the difficulty of acquiring reasonably unfolded UV. Despite recent text-to-3D advancements in supervising multi-view renderings using large text-to-image (T2I) models, issues persist with generation speed, text consistency, and texture quality, resulting in data scarcity among existing datasets. We present \textbf{TexDreamer}, the first zero-shot multimodal high-fidelity 3D human texture generation model. Utilizing an efficient texture adaptation finetuning strategy, we adapt large T2I model to a semantic UV structure while preserving its original generalization capability. Leveraging a novel feature translator module, the trained model is capable of generating high-fidelity 3D human textures from either text or image within seconds. Furthermore, we introduce \textbf{A}r\textbf{T}icu\textbf{L}ated hum\textbf{A}n texture\textbf{S} (ATLAS), the largest high-resolution ($1,024 \times 1,024$) 3D human texture dataset which contains 50k high-fidelity textures with text descriptions. 



\end{abstract} 
\section{Introduction}
\label{sec:intro}

\begin{figure}[tb]
    \centering
    \captionsetup{type=figure}
    \includegraphics[width=\textwidth]{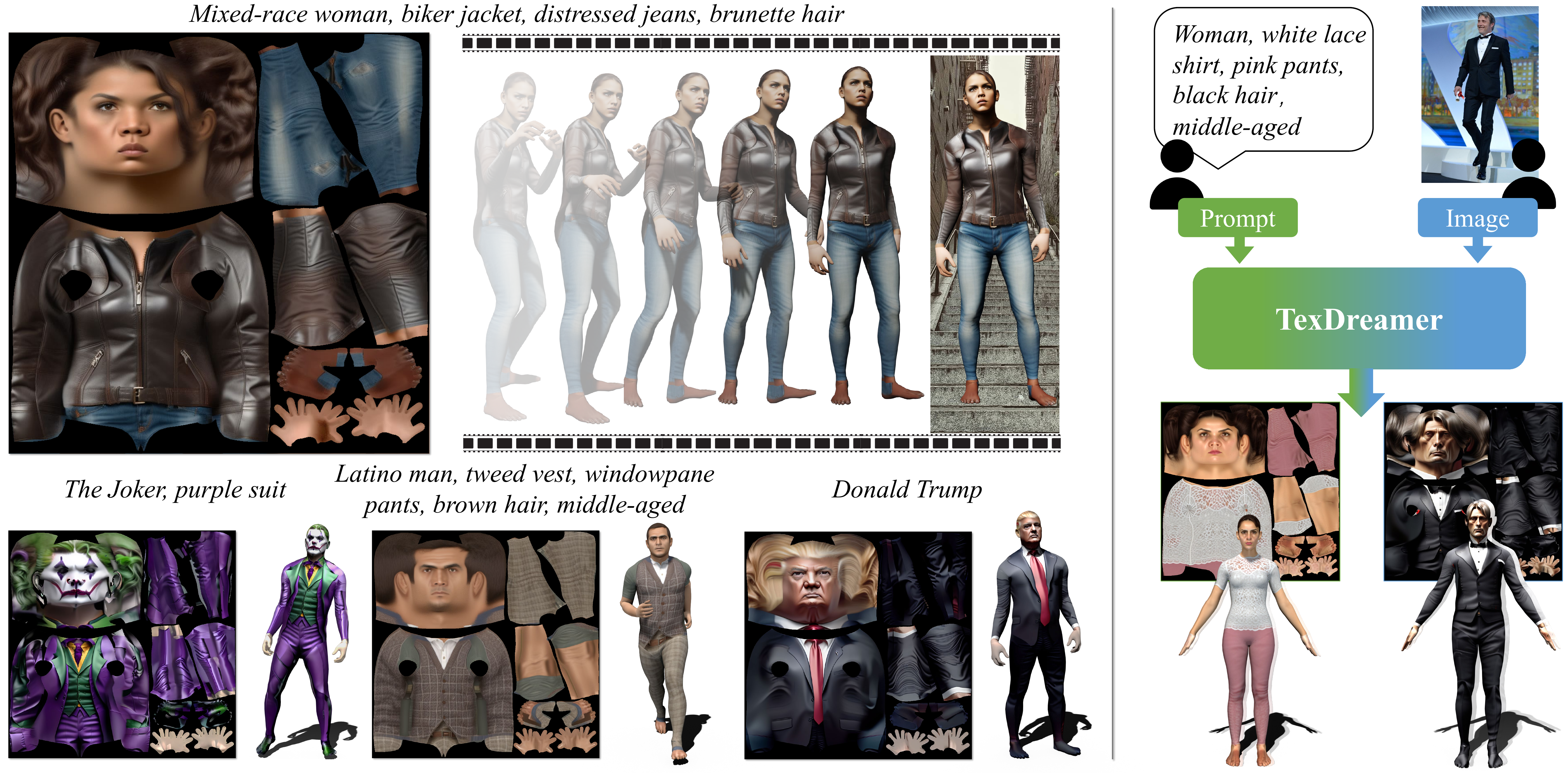} 
    \captionof{figure}{\textbf{Left: Overview of the ATLAS dataset.} ATLAS is so far the largest high-resolution ($1,024\times1,024$) 3D human texture dataset paired with textual descriptions, including both real and fictional identities. \textbf{Right: Basic structure of our TexDreamer.} The first zero-shot high-fidelity human texture generation method that supports both text and image inputs.}
    \vspace{-1.5em}
\label{image:overview}
\end{figure}



3D human texture plays a crucial and essential role in creating appealing 3D human models. UV map allows for seamless and accurate texturing of the 3D model by minimizing distortions, overlapping, and stretching. UV has gained widespread usage in various industrial fields, such as film production, gaming, and virtual reality. 
However, obtaining high-quality textures with reasonably unfolded UV can be a tedious and time-consuming task. In contemporary graphics production, the creation of 3D human textures mainly relies on pricey 3D scanners along with experienced texture painting artists. The scanning process necessitates a capturing system built with multi-camera array and structured light. Texture painting demands the expertise of trained artists proficient in using DCC software, \eg, Substance Painter, ZBrush, and Photoshop. A well-structured human UV map often requires several weeks of dedicated effort. 

Recent significant achievements in text-to-image fields have made it possible to directly generate 3D human models from textual descriptions using 3D priors. However, Human-oriented optimization methods~\cite{avatarbooth,avatarclip,avatarcraft,clip-actor,dreamavatar,dreamhuman,tada} are time-consuming and suffer from limited texture quality due to rendering resolution constraints. Moreover, using these methods in practice requires mesh extraction algorithms such as marching cubes~\cite{marchingcube}, which exhibits challenges in preserving UV layout and mesh topology, making the modification process highly inconvenient. Non-optimization texture generation methods primarily concentrate on objects, approaches like TEXTure~\cite{texture}, Latent-Paint~\cite{latent-paint}, and Text2Tex~\cite{text2tex} focus on completing the multi-view texture of given geometry using the Latent Diffusion Model (LDM)~\cite{LDM}. However, inconsistencies and gaps may occur when dealing with complex input models.

Apart from text, 2D images can also serve as a medium to texture 3D humans. Predicting texture from a single image mainly faces two challenges. For the visible parts, the UV mapping is influenced by the accuracy of pixel-to-surface correspondence estimation. For the invisible parts, the UV results rely on inpainting ability of the model. Without efficient high-quality data, this may lead to artifacts. Video datasets provide multi-view information, which aids in estimating the texture of the invisible parts. However, this approach requires a higher level of precision in pixel-to-surface correspondence estimation across frames. Additionally, video datasets are often limited in quantity. 


To address these issues, we introduce TexDreamer, the first zero-shot multimodal high-fidelity human texture generation method that bridges the gap for 3D human texture creation. Our method can handle two of the most readily available raw data, text and image. This versatility makes our method more flexible and adaptable to different use cases. 
We first conduct an efficient texture adaptation finetuning for our Text-to-UV (T2UV). Training with high-quality sample textures acquired by a novel two-stage texture projection process, T2UV can attend to semantic and positional information of the specific UV structure while preserving the generalization capability of the original T2I model.
For our Image-to-UV (I2UV), instead of predicting invisible parts of partial texture extracted by DensePose~\cite{densepose}, we aim to connect image and UV in a more semantic latent space. We build a feature translator to translate the visual features extracted from the image to the textual feature space of T2UV. Trained with 4.2 million both real and synthetic human images, I2UV shows the highest texture quality and text consistency.
Furthermore, we propose ATLAS (ArTicuLated humAn textureS) dataset, the largest high-resolution ($1,024\times1,024$) 3D human texture dataset. ATLAS contains 50k high-fidelity human textures conforming to the SMPL UV space. Each texture is paired with a detailed text description. See some examples in~\cref{image:overview}, our ALTAS dataset is distinguished by its high-fidelity and diverse character identities.



Our contributions can be summarized as follows: 
\begin{itemize}
\item We introduce TexDreamer, the first zero-shot multimodal high-fidelity 3D human texture generation method accomplished through our efficient texture adaptation finetuning strategy and feature translator module design.
\item We propose ATLAS, the largest high-resolution 3D human texture dataset, filling the vacancy in high-fidelity 3D human texture.
\item Extensive experiments demonstrate that our method surpasses existing approaches regarding text consistency and UV quality for both modalities.
\end{itemize}


\section{Related Work}
\label{sec:related}

\label{sec: 3d human datasets}
\begin{table}[tb]
  \centering \footnotesize
  \caption{Comparisons of our ATLAS with existing human datasets. $^*$ indicates potential acquirable UV textures from 3D scans, and since texture acquirement depends on the setting, their UV resolution remains N/A.  }
  \resizebox{0.9\textwidth}{!}{%
  \begin{tabular}{@{}lccccccc@{}}
    \toprule
    Dataset & 3D Shape & UV Textures & Texture Resolution & Text Description \\
    \midrule
    BUFF~\cite{BUFF} & \ding{51} & $12^*$ & N/A & \ding{55} \\
    CAPE~\cite{CAPE} & \ding{51} & $15^*$ & N/A & \ding{55} \\
    X-Human~\cite{xhuman} & \ding{51} & $20^*$ & N/A & \ding{55} \\
    THuman~\cite{THuman} & \ding{51} & $200^*$ & N/A & \ding{55} \\
    THuman2.0~\cite{THuman2.0} & \ding{51} & $526^*$ & N/A & \ding{55} \\
    Digital Wardrobe~\cite{digitalwordrobe} & \ding{51} & $256^*$ & N/A & \ding{55}  \\
    \midrule
    iPER~\cite{iPER} & \ding{55} & \ding{55} & N/A & \ding{55} \\
    People-Snapshot~\cite{peoplesnapshot} & \ding{51} & 24 & 1,000$\times$1,000 & \ding{55} \\
    SelfRecon~\cite{selfrecon} & \ding{55} & \ding{55} & N/A & \ding{55}\\
    \midrule
    SMPLitex~\cite{smplitex} & \ding{55} & 100 & 512$\times$512 & \ding{51} \\
    SURREAL~\cite{surreal} & \ding{51} & 921 & 512$\times$512 & \ding{55}\\
    \midrule
    \textbf{ATLAS (Ours)} & \textbf{\ding{51}} & \textbf{50k} & \textbf{1,024$\times$1,024} & \textbf{\ding{51}} \\
    \bottomrule
  \end{tabular}
  }
  \label{tab: dataset compare}
  \vspace{-1.5em}
\end{table}

\noindent\textbf{Human-Related Datasets.} 
Comparison of our ATLAS with existing human-related datasets is shown in~\cref{tab: dataset compare}.
3D scans have the highest precision but are the most difficult and time-consuming to acquire.~\cite{BUFF} uses a custom-built multi-camera active stereo system to capture full-body human scans.~\cite{THuman} build THuman with dense DLSR rig, its subsequent~\cite{THuman2.0} provides 500 scans with higher resolution. To predict clothing separately, there are also garment datasets~\cite{CAPE,digitalwordrobe,THuman3.0, synbody}. However, scan data usually has many vertices (often millions) and unstructured grids. Without additional processes, it's hard to obtain texture maps from scans.
In order to be free from complex hardware and high prices, a series of studies~\cite{digitalwordrobe,iPER,peoplesnapshot,selfrecon,video-cap} have proven that neural networks can directly reconstruct 3D human from monocular RGB videos with 3D priors, \eg~the parametric human body model SMPL~\cite{smpl}. Human video datasets~\cite{iPER,peoplesnapshot,selfrecon} generally appear as real human A-pose rotating videos. These kinds of data normally do not include any 3D information.
While some work directly animates 2D image~\cite{zhang2020freenet,xu2023high}, to enhance asset usability and efficiency, many works focus on reconstructing 3D human from a single image~\cite{econ,pamir,pifuhd,tech,makeit3d} leveraging datasets~\cite{deepfashion,market1501,text2human}. Caused by acquisition difficulty, only a few datasets~\cite{360,smplitex,surreal,peoplesnapshot} include UV textures. 
Registering scans, SURREAL~\cite{surreal} provides 921 UV textures. However, on account of the privacy policy, SURREAL UV textures all have the same average face. Lazova~\etal~\cite{360} acquire scans using equipment from~\cite{treedys,twindom} and purchases from commercial datasets~\cite{AXYZ,renderpeople} to contribute a texture dataset, which is laborious and expensive.
Closer to our texture generation method is SMPLitex~\cite{smplitex}, they use 10 UV textures from~\cite{peoplesnapshot,360} to fine-tune Stable Diffusion model~\cite{LDM}, providing 100 UV textures with textual description. However, it lacks variation in identities and clothing. To the best of our knowledge, none of the existing human datasets has the same high-quality textures and rich information as ours.

\noindent\textbf{Texture Generation from Text.}
Significant advancements in text-to-image have drawn considerable attention in the field of text-to-3D generation~\cite{clipmesh,dreamfield,dreamfusion,fantasia3d,magic3d}. Human-oriented optimization methods~\cite{avatarbooth,avatarclip,avatarcraft,dreamavatar,dreamhuman,tada} with SMPL prior show great potential for generating 3D avatars with text. 
AvatarCLIP~\cite{avatarclip} refines the mesh appearance using~\cite{neus} with CLIP score~\cite{clip} supervising on rendering image. Leveraging Score Distillation Score (SDS) loss from DreamFusion~\cite{dreamfusion}, 
AvatarCraft~\cite{avatarcraft} uses NeuS~\cite{neus} combined with Instant-NGP~\cite{instant-ngp} to optimize in canonical space. 
Zero-shot inference methods~\cite{fantasia3d,latent-paint,text2tex,texture,text2mesh} show great advances in texturing 3D objects. Using PBR material~\cite{PBR}, Fantasia3D~\cite{fantasia3d} achieves realistic appearance modeling. Latent-NeRF~\cite{latent-paint} deploys SDS loss in the latent space of LDM~\cite{LDM}. TEXTure~\cite{texture} and Text2Tex~\cite{text2tex} update the multiple viewpoints and inpaint on the 3D mesh texture. Using only 10 training data, SMPLitex~\cite{smplitex} lacks generalized ability and may produce faulty textures.


\noindent\textbf{Texture Generation from Image. }
A group of work~\cite{GTFHA,HPBTT,smplitex,stylepeople,texformer} dedicates to direct texture generation. Image-to-image generation approach~\cite{GTFHA,i2i_1,i2i_2} usually employs a GAN-based network to generate UV. 
Texformer~\cite{texformer} uses a transformer-based network to align 2D human body segmentation with SMPL UV texture. Zhao~\etal~\cite{HPBTT} adds part-based segmentation and enforces cross-view consistency. Stylepeople~\cite{stylepeople} introduces decoupled latent space of GAN to reconstruct hidden parts, but due to the imperfect generation model, it often produces unreasonable results. 
Based on diffusion models, SMPLitex~\cite{smplitex} uses partial segmentation of Densepose as a condition to guide the stable diffusion model.
Another line of work~\cite{360, dinar} considers this task as an inpainting problem. 
Based on DensePose partial segmentation and texture,~\cite{360} uses a GAN-based network to complete texture map and displacement map. DINAR~\cite{dinar} uses StyleGAN2 to convert the input image into a neural texture.
Not like existing methods that rely on 2D image segmentation, we build a feature translator to align human image and UV texture features in latent space.


\section{ATLAS Dataset} 
\label{sec:altas dataset}
\begin{figure*}[t]
  \centering
   \includegraphics[width=1.0\linewidth]{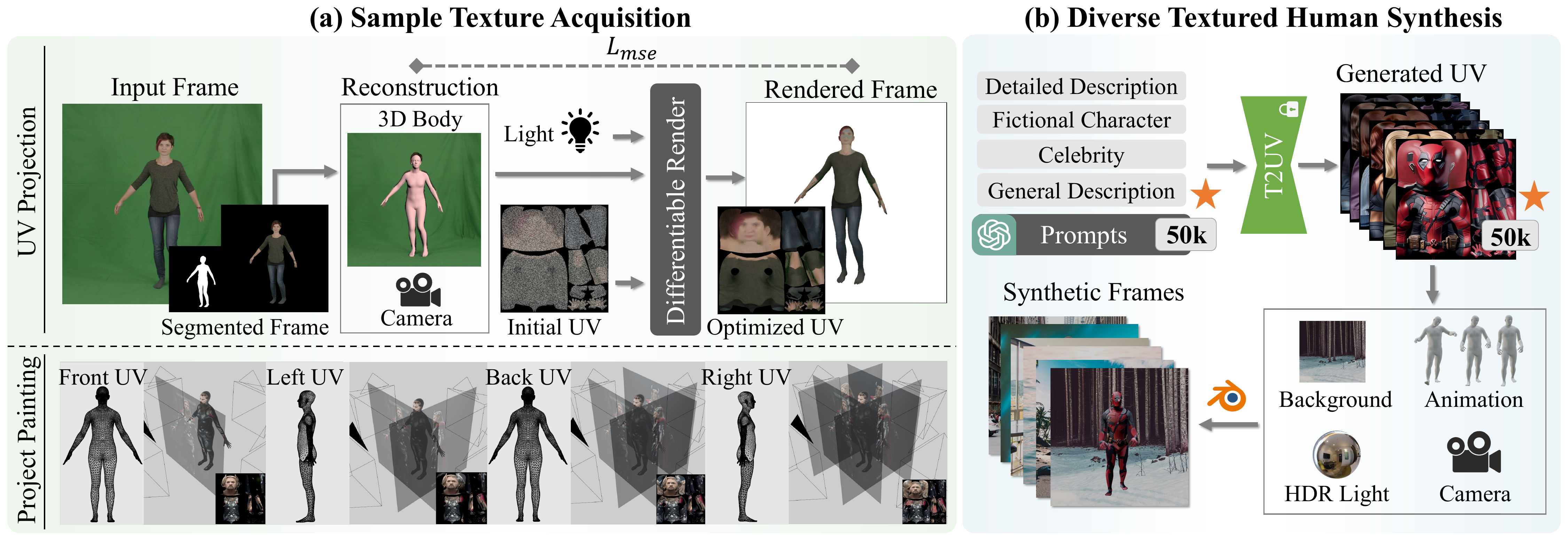}
   \caption{Pipeline for generating synthetic data. Left: Sample texture acquisition. We first use a differentiable render to optimize UV from multi-view images, then further refine them by projection painting. Acquired sample textures with prompts are used to train T2UV in TexDreamer. Right: Diverse textured human synthesis. With the help of ChatGPT, we utilize T2UV to generate 50k human textures. Human images are rendered with animation sequence, background image, HDR lighting, and perspective camera. Orange stars indicate included data in our ATLAS dataset.}
   \label{image: pipeline}
   \vspace{-1.5em}
\end{figure*}

Producing large-scale human textures with reasonably unfolded UV is inherently challenging due to the difficulty of acquiring such data for training. This section presents our ArTicuLated humAn textureS (ATLAS) dataset and describes its data generation strategy for TexDreamer training, including sample textures acquistion~\cref{subsec: sample texture acquisition} and diverse textured human synthesis~\cref{subsec: Diverse Textured Human Synthesis}. See ATLAS visual pipeline in~\cref{image: pipeline}.

\subsection{Sample Texture Acquistion}
\label{subsec: sample texture acquisition}

Obtaining well-structured human UV textures traditionally needs to register from scan data or paint by artists. We surpass both and propose first use UV Projection to optimize coarse human UV from multi-view images and then refine them with project painting. See the left of~\cref{image: pipeline} for their visual process. 

The core idea of UV projection is to minimize the difference between ground-truth frames and rendered frames. After segmentation, We exploit CLIFF~\cite{cliff} to estimate global rotation, joint pose, and 3D shape, along with camera parameters from masked frames. The initial UV map can be optimized through differentiable rendering. However, deviations exist between the estimated pose and the actual pose. We further conduct Project Painting to improve UV quality, which is a texture painting technique commonly used in CGI production. UV texture quality can be improved by alternating and modifying different SMPL UV maps from multiple view angles. The obtained UV data is used to train TexDreamer T2UV, see its detailed training method in~\cref{subsec: t2uv}. 

To increase sample texture diversity and avoid T2UV overfitting, we use both real and generated multi-view images. For real human texture, we use videos from People-Snapshot~\cite{peoplesnapshot} and iPER~\cite{iPER}. As for fictional characters, we rely on ControlNet~\cite{controlnet} and DWpose~\cite{DWpose} with pretrained LDM~\cite{LDM} to generate multi-view images of desired identities. Using eight SMPL A-pose for each character (rotation angle: $0, \pm 45, \pm90, \pm135, 180$) and textual reinforcement, we manage to deal with the ID consistency problem of LDM. Specifically, we add corresponding orientation descriptions to constrain the generation, both positively and negatively. For instance, regarding the backside generation, we use ``the back of, backside'' as positive prompt, and the corresponding negative prompt is ``face, front''.


\subsection{Diverse Textured Human Synthesis}\label{subsec: Diverse Textured Human Synthesis}

To synthesize diverse textured human images with identities for I2UV training, we composite T2UV-generated textures with animation, background, and HDR lighting. See the right of~\cref{image: pipeline} for a visual pipeline.  \\
\noindent\textbf{Texture Generation.} Generating a UV dataset using T2UV requires a large number of corresponding text descriptions. Expanding AvatarCLIP~\cite{avatarclip} classification, we depict our description as four categories: detailed description, fictional character, celebrity, and general description. Each has a designed structure. For detailed description, we first describe the appearance of a person with randomized descriptions for race or country, followed by gender, clothing, hairstyle, and age. For fictional characters and celebrities, we depict the person's name and their common clothing. Celebrities add hairstyles to the structure of fictional characters. As for general description, each prompt contains one word or phrase to represent the category. See more prompt design in the supplementary. Leveraging ChatGPT~\cite{chatgpt}, we generated a total of 50k prompts. We randomly select 20\% of generation as ATLAS test set. \\
\noindent\textbf{Composite Rendering.} To enhance authenticity, we synthesize human images using Blender~\cite{blender} with HDR image lighting and PBR human material shaders~\cite{PBR}. HDR lighting is developed by the Image Based Lighting (IBL) process~\cite{IBL}, in which the light is sampled from a 360\degree~panoramic image and reused to relight entire CG scene. IBL can simulate real scenes and ensure uniform lighting, we use HDR images as the ``sunlight'' to enrich lighting. For human material, we use bidirectional reflectance distribution function (BSDF), which is a variant from Disney principled model also known as PBR shader~\cite{PBR}. See detailed settings in the supplementary.

Diversification of human postures is accomplished by AMASS~\cite{amass}, the largest human motion capture dataset, including more than 40 hours of motion data, spanning over 300 subjects. All the motion rate is set to 24, equal to the render frame rate, producing over 8.3 million rendering frames. Every motion sequence has a global transformation. To capture each human motion more completely, we set constraints on the rendering camera. Each perspective camera tracks the movement of ``pelvis'' joint and is located 5 meters in front of the mesh with 80mm focal length. The rendering sample for each pixel is set to 64. 
 
Incorporating backgrounds contributes to a closer resemblance of real human images in the wild and increases the richness of synthetic images. Previous synthetic dataset SURREAL uses categories of kitchen, living room, bedroom, and dining room from LSUN~\cite{lsun}. LSUN image resolution is very low ($256\times256$). To increase realism, we use royalty-free images from Pexels~\cite{pexel}, including natural scenes, urban streets, indoor settings, abstract textures, and plain colors. Applying post-processing technique to compute input ``alpha'' channels and layer multiple rendering channels, we can combine the textured motion sequences with background images.
\section{Zero-Shot Human Texture Generation} 
\label{sec:texdreamer}


Generating large-scale realistic human textures with a uniform and semantic UV layout is inherently challenging due to the difficulty of acquiring efficient training data. We aim to use a small number of sample textures and leverage the generative and generalization capabilities of pretrained large-scale T2I models to establish a connection between common character generation and corresponding UV components.
In this section, we provide a detailed description of TexDreamer, the first zero-shot multimodal high-fidelity 3D human texture generation method. We conduct a two-step training stradegy, Text-to-UV (T2UV)~\cref{subsec: t2uv} and Image-to-UV (I2UV)~\cref{subsec: i2uv}. We first train the T2UV module with efficient texture adaptation fine-tuning to allow texture generation from text. Then utilizing T2UV along with synthesized rendering images from ATLAS data generation, we train I2UV using a novel feature translator.

\subsection{Preliminaries} \label{fine_tune_sd}

Dreambooth~\cite{dreambooth} finetunes the entire parameters in LDM and creates a new checkpoint. Dreambooth can yield impressive results, but they come at a cost in terms of size. Textual Inversion~\cite{text-inversion}, on the other hand, is faster because they learn to represent the provided concept through new ``words'' in the text embedding space. However, they only work for a single or a small handful of subjects. Different from the above two methods, low-rank adaption (LoRA)~\cite{lora} method adds a new set of weights to the model, which can be used for general-purpose controlling. LoRA is initially proposed to fine-tune large language models (LLMs), it learns weights by adding extra layers in the transformer cross-attention layer and uses low-rank matrices to learn the offset of parameters. This technique can also be applied in LDM.

\begin{figure}[t]
  \centering
   \includegraphics[width=1.0\linewidth]{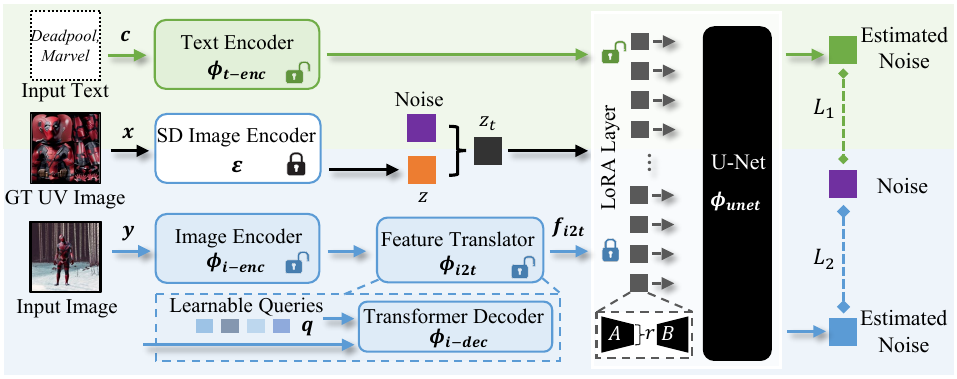}
   \caption{Structure of TexDreamer. We conduct two training stages. For T2UV (green), we use LDM denoise loss $\mathcal{L}_1$ to optimize the text encoder and U-Net. For I2UV (blue), the feature translator $\phi_{i2t}$  map the input image feature encoded by $\phi_{i-enc}$ to a conditional feature $f_{i2t}$. We train I2UV by optimizing $\phi_{t-enc}$ and $\phi_{i-enc}$ with $\mathcal{L}_2$.}
   \label{image: method}
   \vspace{-1.5em}
\end{figure}

\subsection{Text-to-UV}\label{subsec: t2uv} 
For T2UV training, we conduct efficient texture adaptation fine-tuning. See the green flow of~\cref{image: method} for a visual training process of T2UV. Specifically, we add a few trainable parameters in each attention layer and train the model to learn the specific common concept of a small dataset through LoRA fine-tuning. Among all the finetune methods~\cref{fine_tune_sd}, LoRA strikes a good balance between training efficiency and the ability to fine-tune the model to generate specific concepts. Typically, the weight matrices in dense layers have full rank. LoRA shows that the updates to the weights also have a low ``intrinsic rank'' during adaptation. For the pre-trained weight matrix of LDM $W_{\phi_{unet}} \in \mathbb{R}^{d \times k}$, LoRA constrain the update by representing the latter with a low-rank decomposition, in this case $W_{\phi_{unet}} + \Delta W = W_{\phi_{unet}} + BA$, where $B \in \mathbb{R}^{d \times r}, A \in \mathbb{R}^{r \times k}$, and the rank $r \ll \min(d,k)$. During training, the trainable parameters are in $A$ and $B$. For image-text input latent $s$, $\tilde{s} = W_{\phi_{tenc-unet}} s$, the modified forward pass yields:

\begin{align}
    \label{eq: lora}
    \scalebox{0.90}{$\tilde{s} =W_{\phi_{tenc-unet}} s+\Delta W s=W_{\phi_{tenc-unet}} s+B A s.$}
\end{align}

LoRA uses a random Gaussian initialization for A and zero for B, then scales $W_{\phi_{tenc-unet}} s$ by $\frac{\alpha}{r}$, where $\alpha$ is a constant in $r$. The input GT UV image $x$ is encoded with SD image encoder $\mathcal{E}$ and the input text is encoded by $\phi_{t-enc}$. Due to the large number of hyper-parameters of LDM, there is no fixed universal configuration during training. Using sample texture acquired from~\cref{subsec: sample texture acquisition} together with their prompts $c$, we train T2UV via:
\begin{equation}
    \label{eq:t2uv_ldm}
    \scalebox{0.90}{$L_{1}:=\mathbb{E}_{\mathcal{E}(x), c, \epsilon \sim \mathcal{N}(0,1), t}\left[\left\|\epsilon-\phi_{unet}\left(z_t, t, \phi_{t-enc}(c)\right)\right\|_2^2\right],$}
\end{equation}   
where both text encoder $\phi_{t-enc}$ and U-Net $\phi_{unet}$ are jointly optimized by~\cref{eq:t2uv_ldm}. 

With appropriate scale initialization, tuning $\alpha$ and $r$ is nearly the same as tuning the learning rate, slight change may result in evident differences. In order to find the best $\alpha$ and $r$ for optimizing $\phi_{t-enc}$ and $\phi_{unet}$, other than $\mathcal{L}_{1}$ value, we mainly depend on quantitative measurement CLIP score~\cite{clip}. For each setting, we calculate CLIP score on rendered T pose images and the corresponding prompts. To enhance the text-image consistency, we further employ an alignment enhancement strategy. After training with sample textures, we use the trained T2UV to generate 4 textures and only select 1 texture with the highest CLIP score for each prompt in ATLAS. The final T2UV we use for I2UV has the best text consistency, see~\cref{subsec: ablation} for T2UV ablation study. Moreover, the number of training samples can also influence model capabilities, see more experiments regarding this in the supplementary.
 \begin{figure}[t]
  \centering
   \includegraphics[width=1.0\linewidth]{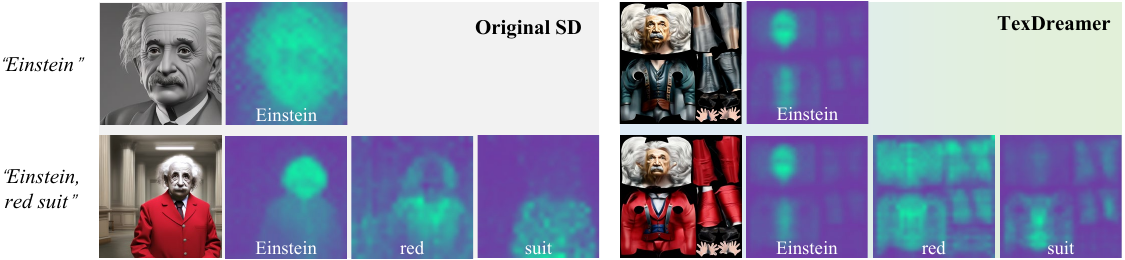}
   \caption{Comparison of attention maps between the original SD and TexDreamer T2UV. The response area of the original SD is random, while T2UV consistently maps the prompts to the learned UV structure.}
   \label{image: attention}
   \vspace{-1.5em}
\end{figure}

After extensive experiments, we find that T2UV inherits the generalization capabilities of the original SD model while adapting generation to a semantic UV layout. With the same text input, the attention response area of the original SD is random, while TexDreamer T2UV consistently maps the texts to the learned UV structure, see comparisons in~\cref{image: attention}. This learned structural information indicates that TexDreamer has the potential to generate a large-scale of human textures with various identities and clothing.


\subsection{Image-to-UV}\label{subsec: i2uv}
For predicting invisible textures from 2D human images, our key insight is that different structures of human images and UV textures can be connected with more semantic medium. In this case, we use the textual feature.

Training strategy of I2UV depends on synthetic textured human images from ATLAS and previously trained T2UV, see the blue flow in~\cref{image: method} for a visual process. Inspired by~\cite{minigpt}, we build a novel feature translator to transform input 2D image feature to the conditional text feature space of T2UV. 
We use image encoder $\phi_{i-enc}$ from CLIP~\cite{clip} to extract visual features $f_{voken}$ from rendered 2D image $y$. The feature translator translates visual features to textual features $f_{i2t}$, which includes a two-layer MLP model $\phi_{MLP}$, a three-layer of transformer decoder $\phi_{i-dec}$, and a learnable query sequence $q$. The translated feature $f_{i2t} \in \mathbb{R}^{L \times \hat{d}}$ is then formulated as: 
\begin{align}
    \label{eq:i2uv f_i2t}
    f_{i2t} :=\phi_{i-dec}\left(\phi_{MLP}\left(f_{voken}\right), q\right) \in \mathbf{R}^{L \times \hat{d}},
\end{align}
where $L$ is the maximum input length of text encoder $\phi_{t-enc}$ and $\hat{d}$ is the dimension of LDM encoder $\mathcal{E}$ output feature. In our case, $f_{i2t} \in \mathbb{R}^{77 \times 1,024}$. 


In order to constrain T2UV generation with the input image, the mapped feature $f_{i2t}$ functions as a condition in the generation process. When training, the generated UV texture, as the ground truth, is first encoded into a latent feature $z_0$ through LDM image encoder. The noisy feature $z_t$ is obtained by adding noise $\epsilon$ to $z_0$. As $f_{i2t}$ is essentially a text feature, similar to $\phi_{t-enc}(c)$, it can be directly used in training, we optimize the image encoder $\phi_{i-enc}$ and feature translator $\phi_{i2t}$ with LDM denoise loss:
\begin{equation}
    \label{eq:i2uv_ldm}
    \begin{aligned}
        &f_{i2t} :=\phi_{i2t}(\phi_{i-enc}(y)), \\
        &L_{2}:=\mathbb{E}_{\mathcal{E}(x), y, \epsilon \sim\mathcal{N}(0,1), t}\left[\left\|\epsilon-\phi_{unet}\left(z_t,t, f_{i2t}\right)\right\|_2^2\right].
    \end{aligned}
\end{equation}

\section{Experiments} 
\label{sec:experiment}
\subsection{Experimental Setup}
\label{subsec: experimental setup}

\noindent\textbf{Implementation Details.} For training T2UV, we use stable-diffusion-2-1 and clip-vit-large-patch14-336. The rank and $\alpha$ for $\phi_{unet}$ is 128, for $\phi_{t-enc}$ is 16. Each training uses batch size 8, with a total of 2,000 training steps. The optimizer is AdamW, with a learning rate set to 0.0001 and a constant scheduler with 100 warm-up steps. To improve training efficiency, we followed~\cite{snr} and set the SNR-$\gamma$ to 5. During inference, the weight of T2UV is set to 1.0, and we use the results of 32 steps. For training I2UV, based on T2UV, we use the same batch size but add training steps to 20,000, change the learning rate to $1e-5$, and set weight decay in regularization at 0.01. All training is conducted on a single Nvidia A100 GPU.

\noindent\textbf{Evaluation Metrics}. For T2UV, we use CLIP score~\cite{clip} to measure consistency between generated textures and input texts. For all calculations, we render generated textures with SMPL neutral body in T-pose using Pytorch3D with the same perspective camera, lighting, and material. For I2UV, previous methods~\cite{HPBTT,RSTG,TexGlo,texformer} mainly use SSIM~\cite{SSIM} and LPIPS~\cite{LPIPS} compute between the renderings and ground-truth images. However, affected by the accuracy of human pose estimation, these metrics can not fully measure the reconstructed texture quality. Since we have texture ground truth and corresponding text, we propose to use Mean Squared Error (MSE) and CLIP score to evaluate both texture quality and text consistency.

\begin{figure}[t]
  \centering
   \includegraphics[width=1.0\linewidth]{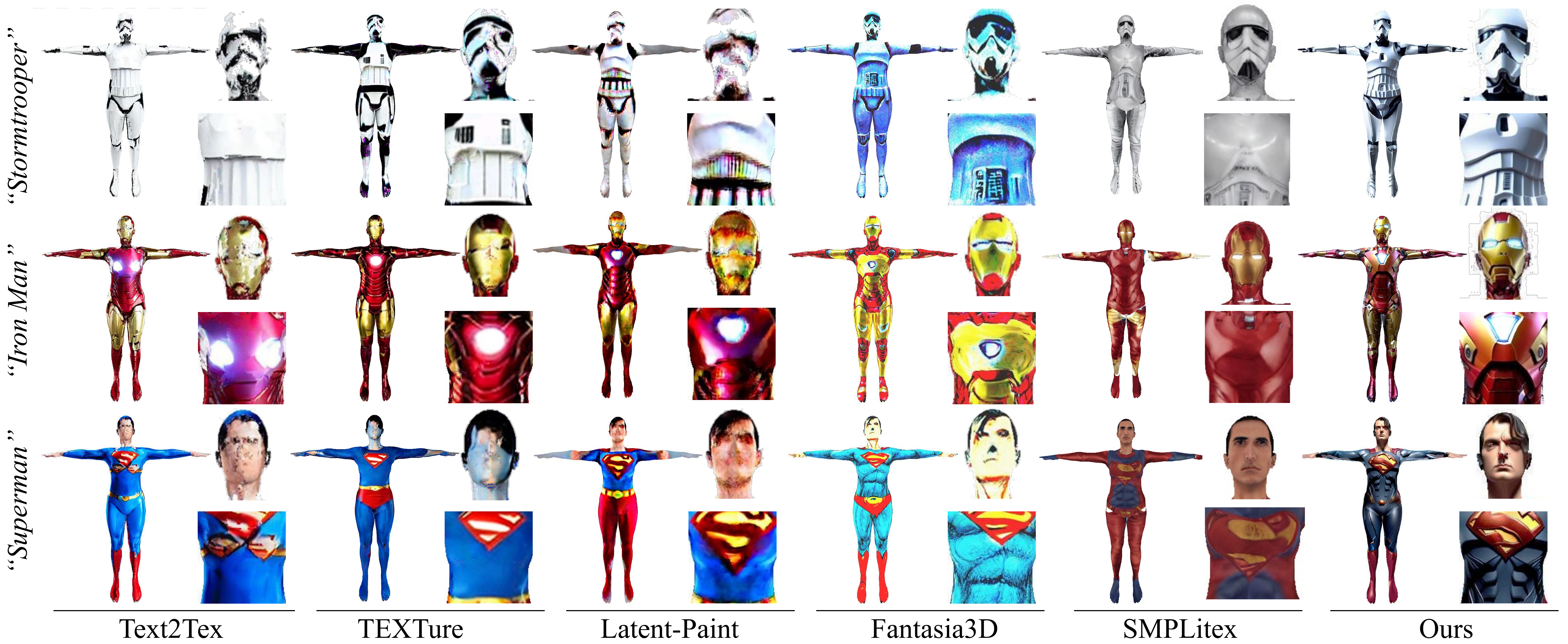}
   \caption{Qualitative comparison of texture generation from text. We compare TexDreamer with state-of-the-art texture generation methods, including Text2Tex\cite{text2tex}, TEXTure~\cite{texture}, Latent-Paint~\cite{latent-paint} and Fantasia3D~\cite{fantasia3d}. Our results clearly achieve the finest facial details and the highest overall quality. Please zoom in for a better view.}
   \label{image: t2uv compare}
   \vspace{-0.7em}
\end{figure}

\begin{figure}[t]
  \centering
   \includegraphics[width=0.9\linewidth]{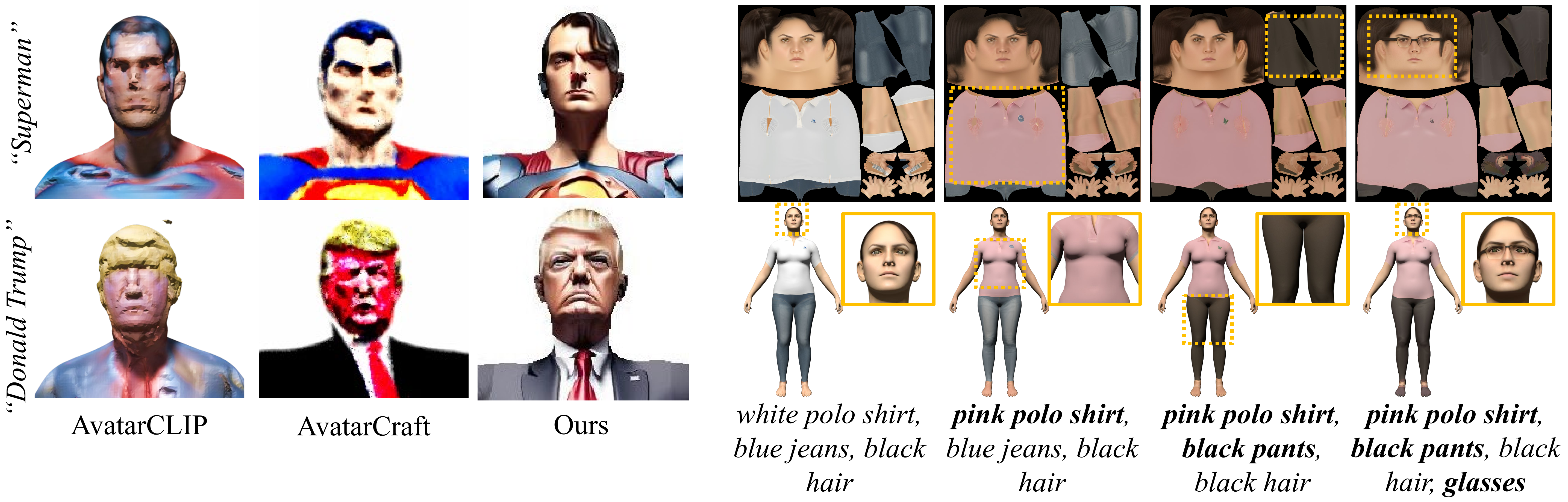}
   \caption{Left: Qualitative comparison with AvatarCLIP~\cite{avatarclip} and AvatarCraft~\cite{avatarcraft}. Our method has more realistic head avatars. Right: Texture editing. TexDreamer can use text to edit generated texture details,~\eg, clothing style, color, and accessories.}
   \label{image: t2uv compare head-detail}
   \vspace{-1.5em}
\end{figure}

\subsection{Qualitative Comparison}
\label{subsec: qualitative comparison}
\noindent\textbf{T2UV.} We compare TexDreamer T2UV with state-of-the-art texture generation methods, including Text2Tex\cite{text2tex}, TEXTure~\cite{texture}, Latent-Paint~\cite{latent-paint}, Fantasia3D~\cite{fantasia3d}, and SMPLitex~\cite{smplitex}. 
As shown in~\cref{image: t2uv compare}, our generation achieves the highest overall quality and the finest facial details. Since human-oriented optimization methods require a long time, see~\cref{tab: t2uv quantity}, we choose the first~\cite{avatarclip} and most current advanced~\cite{avatarcraft} open source method to compare. They both have additional optimization steps in the facial area. We use identities in their showcase, left of~\cref{image: t2uv compare head-detail} shows their results lack realism in texture colors and facial features. See more generation results of TexDreamer in the supplementary.

\begin{figure}[t]
  \centering
   \includegraphics[width=1.0\linewidth]{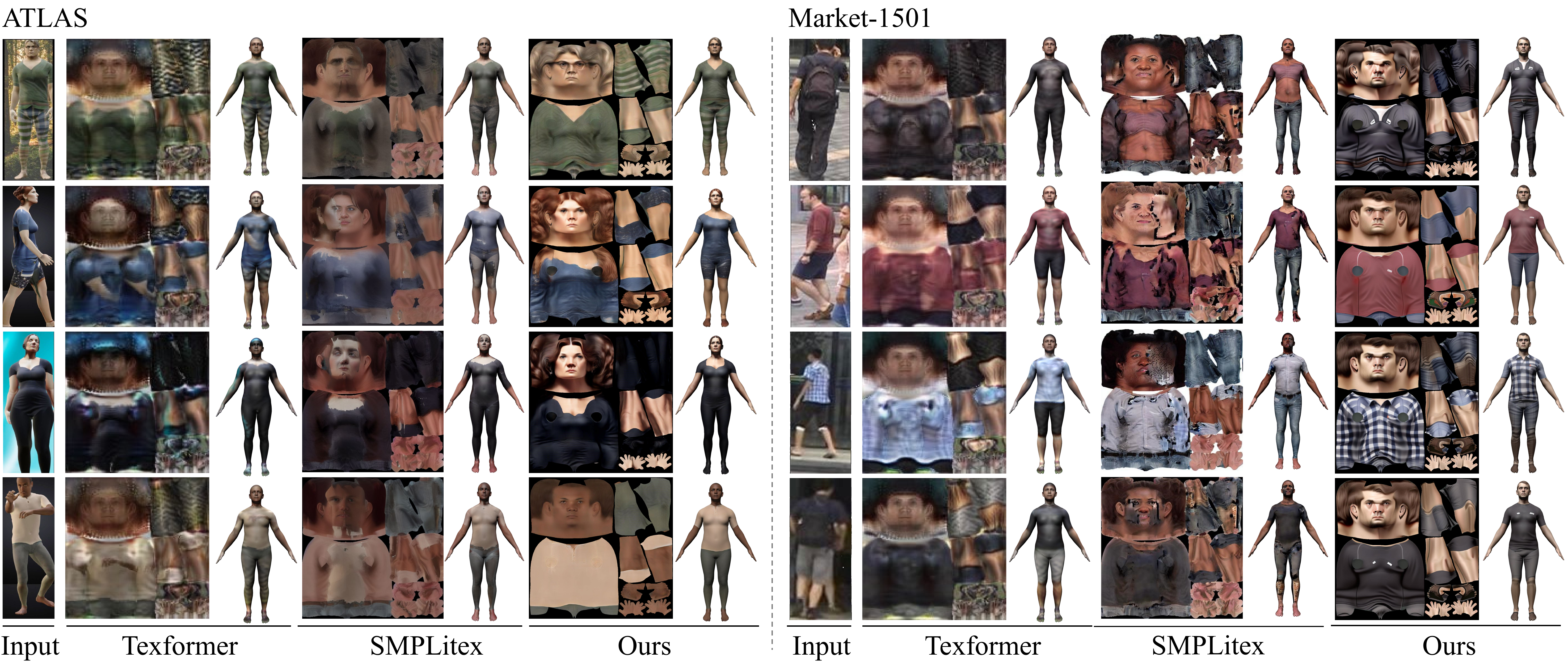}
   \caption{Qualitative comparison with UV generation from image. We compare with advanced Texformer~\cite{texformer} and SMPLitex~\cite{smplitex} on our ATLAS dataset (Left) and Market-1501~\cite{market1501} (Right). Please zoom in to compare texture completeness and quality.}
   \label{image: i2uv compare}
   \vspace{-1.5em}
\end{figure}

\noindent\textbf{I2UV.} We compare image-to-uv with the leading method Texformer~\cite{texformer} and SMPLitex~\cite{smplitex}. We evaluate both on Texformer training dataset Market-1501~\cite{market1501} and our ATLAS test set. See visual comparisons in~\cref{image: i2uv compare}. Our method achieves remarkably faithful identities and outstanding texture realism.

\subsection{Quantitative Comparison}

\noindent\textbf{T2UV.} The rendering of textured 3D human from text should closely resemble the input text at the reference view, and demonstrate consistent semantics with the reference under novel views. We evaluate these two aspects with CLIP score~\cite{clip}, which computes the semantic similarity between the novel view and the reference. To evaluate T2UV text consistency more comprehensively, in addition to the basic rendering of SMPL in~\cref{subsec: experimental setup}, we add three more views (azim = 0, 90, 180, 270). Due to the high consumption of time and resources, we only compare efficiency with AvatarCraft.
As shown in~\cref{tab: t2uv quantity}, our method is the most efficient and achieves the best text consistency. 

\noindent\textbf{I2UV.} With generated and collected textures as ground truth, we first use MSE to compare I2UV with advanced image-to-UV method Texformer~\cite{texformer} and Smplitex~\cite{smplitex}. MSE computes the average of the squares between predicted and actual values. The smaller the MSE, the higher texture generation quality of the model. We randomly extract two frames from ATLAS test set as input. Furthermore, we evaluated text consistency of each texture using CLIP score with paired rendering and text description.~\cref{tab: i2uv quantity } shows TexDreamer I2UV achieves the best result on both measurements.

\begin{table}[b]
    \centering \footnotesize
    \vspace{-2.0em}
    \caption{Quantitative comparison of generating human texture from text . ``-T'' means TexDreamer T2UV, we show compared inference methods (up) and optimization methods (middle) results. CLIP score of AvatarCraft is not reported due to the high consumption of time and resources.}
    \resizebox{0.6\textwidth}{!}{%
    \begin{tabular}{lccc}
    \toprule
    Method & GPU (GiB) & Time (mins) $\downarrow$ & CLIP Score $\uparrow$ \\
    \midrule
    Text2Tex~\cite{text2tex} & 20.31 & $\sim$ 14.35 & 29.962  \\
    TEXTure~\cite{texture} & 12.05 & $\sim$ 2.38 & 27.298\\
    Latent-Paint~\cite{latent-paint} & 11.46 & $\sim$ 13.95 & 26.378 \\
    Fantasia3d~\cite{fantasia3d} & 12.42 & $\sim$ 14.50 & 30.557  \\ 
    SMPLitex~\cite{smplitex} & 7.77 & $\sim$ 0.31 & 22.998 \\
    \midrule
    AvatarCLIP~\cite{avatarclip} & 37.74 & $\sim$ 360 & 29.422 \\
    AvatarCraft$^*$~\cite{avatarcraft} & 26.65 & $\sim$ 480 & -\\
    \midrule
    \textbf{Ours-T2UV} & \textbf{5.71} & $\sim$ \textbf{0.17} & \textbf{31.297}\\
    \bottomrule
  \end{tabular}
  }
  \label{tab: t2uv quantity}
\end{table}

\noindent\textbf{User Study.} 
We further conduct a user study to evaluate texturing 3D humans using text.~\cref{tab: t2uv user study} indicates that our method has the highest preference and resembles closest to corresponding text. We use the same rendering view images, and invite 14 participants to rate on a scale of 1-5 about ``the quality of overall texture'' and ``consistency with text description''. Each participant is randomly assigned with the same amount of comparisons.

\begin{table}[t]
  \centering \footnotesize
  \caption{User study on texture generation from text. Our result has the highest image quality and test consistency. }
  \resizebox{0.6\textwidth}{!}{%
  \begin{tabular}{@{}lcc@{}}
    \toprule 
    Method & Text Consistency $\uparrow$ & Image Quality $\uparrow$\\
    \midrule
    Text2Tex~\cite{text2tex} & 1.919 & 1.641 \\
    TEXTure~\cite{texture}	& 2.003 & 1.744 \\
    Latent-Paint~\cite{latent-paint} & 1.878 & 1.456 \\
    Fantasia3D~\cite{fantasia3d} & 2.089 & 1.904 \\
    AvatarCLIP~\cite{avatarclip} & 1.752 & 1.341 \\
    \textbf{TexDreamer (Ours)} & \textbf{4.019} & \textbf{4.244} \\
    \bottomrule
  \end{tabular}
  }
  \label{tab: t2uv user study}
  \vspace{-0.5em}
\end{table}


\begin{table}[t]
    \begin{minipage}[t]{0.45\textwidth}
         \centering
          \caption{Quantity comparison and ablation study of TexDreamer I2UV. ``fixed $\phi_{i-enc}$" means we don‘t train the image encoder in I2UV. }
          \resizebox{1.0\textwidth}{!}{%
          \begin{tabular}{@{}lcc@{}}
            \toprule 
            Method & MSE $\downarrow$ & CLIP Score $\uparrow$\\
            \midrule
            Texformer~\cite{texformer} & 0.1148 & 21.811\\
            SMPLitex~\cite{smplitex} & 0.0783 & 22.488 \\
            Ours-I2UV (fixed $\phi_{i-enc}$) & 0.0632 & 26.138 \\
            \textbf{Ours-I2UV (full)} & \textbf{0.0442} & \textbf{27.334}\\
            \bottomrule
          \end{tabular}
          }
          \label{tab: i2uv quantity }
    \end{minipage}
    \hfill
    \begin{minipage}[t]{0.45\textwidth}
    \centering \footnotesize
       \caption{Ablation study of TexDreamer T2UV. Our setting has the highest text consistency.}
       \resizebox{1.0\textwidth}{!}{%
       \begin{tabular}{cccccc}
        \toprule
        $\phi_{unet} r$ & $\phi_{unet} \alpha$ & $\phi_{t-enc} r$ & $\phi_{t-enc} \alpha$ & CLIP Score $\uparrow$ \\
        \midrule
        128 & 128 & 8 & 8 & 28.64 \\
        \textbf{128} & \textbf{128} & \textbf{16} & \textbf{16} & \textbf{29.29} \\
        128 & 128 & 32 & 32 & 28.36 \\
        \midrule
        64 & 64 & 16 & 16 & 28.20 \\
        192 & 192 & 16 & 16 & 29.19 \\
        \bottomrule
      \end{tabular}
      }
      \label{tab: t2uv ablation}
    \end{minipage}
      \vspace{-1.5em}
\end{table}

\subsection{Ablation Study}
\label{subsec: ablation}

For training texture generation from text, we add a few trainable parameters in each attention layer, a slight change of rank and $\alpha$ in LoRA can greatly impact the generation result. We conducted ablation experiments on $r$ and $\alpha$ of U-Net and text encoder. For ablation purposes, we reduce the training steps in~\cref{subsec: experimental setup} setting, others remain the same. ~\cref{tab: t2uv ablation} shows that our T2UV has the highest text consistency. For I2UV, we compare a fixed text encoder and full I2UV module on ATLAS test set, see \cref{tab: i2uv quantity }. Our full model has the highest image similarity and text consistency.

\section{Applications}
\label{sec:application}


\begin{figure}[t]
  \centering
   \includegraphics[width=0.8\linewidth]{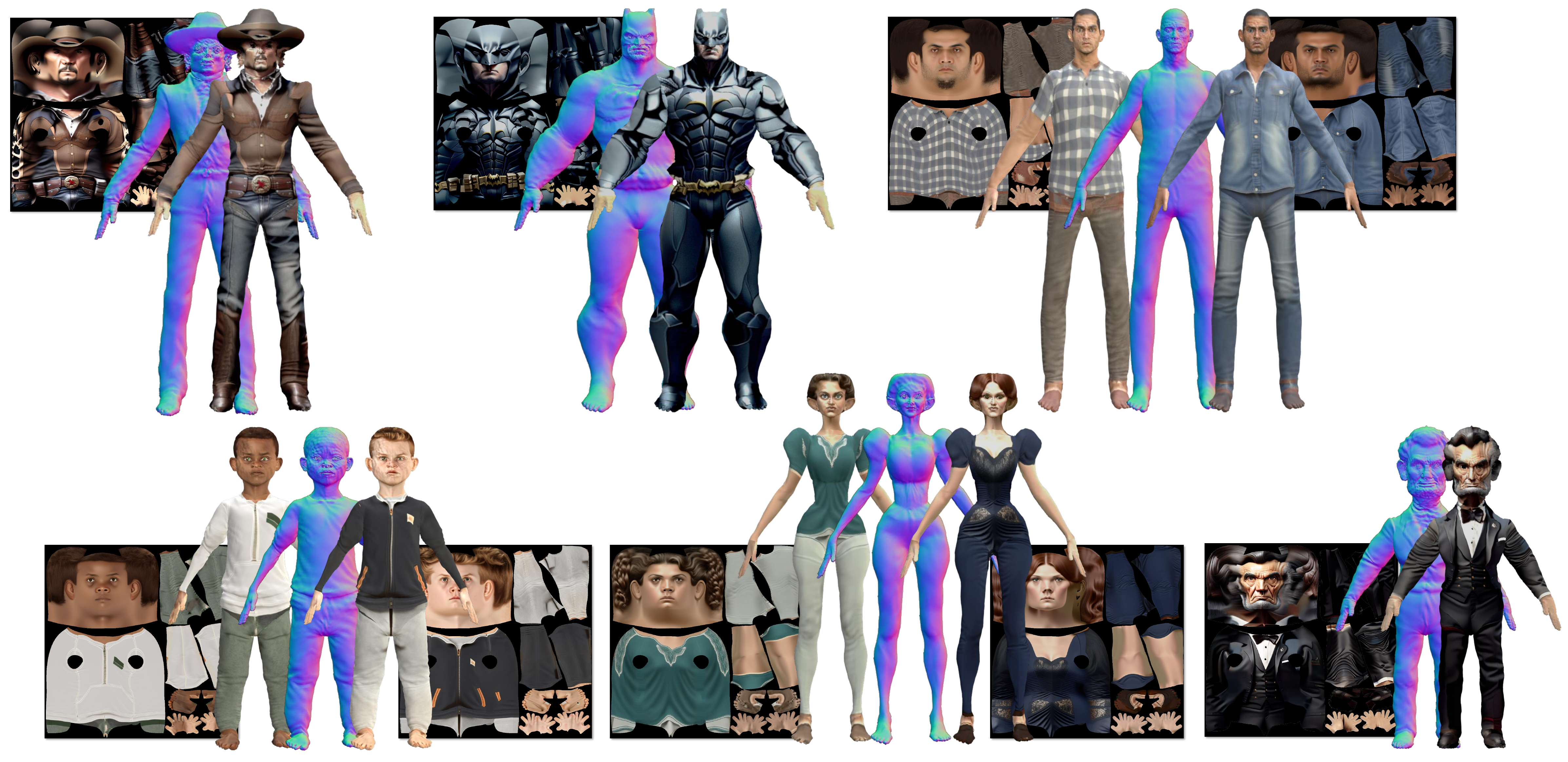}
   \caption{Texturing dressed avatars. Our human textures can be applied to complex dressed meshes generated by text-to-3d method. We show some examples generated by TADA~\cite{tada} with synthetic UV texture generated by TexDreamer.}
   \label{image: virtual try-on}
   \vspace{-2.0em}
\end{figure}

\textbf{Texture Editing.} TexDreamer can use text to edit 3D human appearance generation, details to clothing style (both upper and lower garments), accessories, \etc, see~\cref{image: t2uv compare head-detail}. This allows fast and precise alterations of characters designed by artists, making our method more flexible and adaptable to concept character design in the film or game industry. The visual results also indicate that our method can generate human textures with different clothing while preserving the identity, which additionally opens doors for a fast virtual try-on.

\noindent\textbf{Texturing Dressed Avatars.} We further explored the application of our generated textures for more complex geometry. As shown in~\cref{image: virtual try-on}, our generated textures can integrate with complex human mesh and produce more authentic human-like characters. Specifically, we leverage the most advanced text-to-3D-avatar generation method TADA~\cite{tada}. We apply the synthetic textures from TexDreamer by modifying the mesh initialization process, allowing it to density mesh while preserving the original UV information. This application shows that TexDreamer empowers users to create personalized characters with ease, which could be laborious for traditional 3D modeling techniques.







\section{Conclusions} 
\label{sec:conclusions}
We propose the first zero-shot multimodal high-fidelity 3D human texture generation model, TexDreamer. Adapting large T2I model generative ability to unique UV structure with efficient texture adaptation fine-tuning and a novel feature translator, TexDreamer exhibits faithful identity and clothing for texturing 3D humans using texts or images. Enabling a more diverse range of human-like avatar generation.
Furthermore, we construct ATLAS, the most extensive high-resolution (1,024$\times$1,024) 3D human texture dataset with a uniform and semantic UV layout, filling the absence of high-quality human UV data. Extensive experiments demonstrate that our method surpasses existing approaches in terms of text consistency and UV quality.

\vspace{1mm}
\noindent\textbf{Limitations and Social Impacts.}
While TexDreamer shows promising results, it still has several limitations. 
I2UV is not based on Densepose segmentation, when applied to real-life cases, some output may not strictly align with the input clothing pattern. 
Producing realistic human texture, Texformer has the potential to influence virtual human industries. However, it also raises ethical and privacy concerns, as the technology could potentially be used for creating deepfakes.

%
%
\bibliographystyle{splncs04}
\bibliography{main}

\clearpage
\begin{center}
    \textbf{\Large Appendix}
\end{center}
\appendix



\noindent In this appendix, ~\cref{sec: More Details of ATLAS Dataset} provides more information regarding the ATLAS dataset construction.~\cref{sec: More Analysis} presents more analysis of TexDreamer training.~\cref{sec:More Qualitative Results} show more qualitative results of TexDreamer, including both text-to-UV and image-to-UV. We also present more human textures included in our ATLAS dataset.


\section{More Details of ATLAS Construction}
\label{sec: More Details of ATLAS Dataset}

\noindent\textbf{Text Augmentation.}
To acquire multi-view images for fictional characters, we use text augmentation to promote the consistency of generated character identities. ~\cref{tab: prompts for datasets} provides details about view-related and other description prompts.
The positive prompt $T_{pos}$ is used to condition the main generation direction. We describe $T_{pos}$ with character identity $T_{id}$, generate poses $T_{pose}$ and other descriptions $T_{other}$.

\begin{figure}[t]
  \centering
   \includegraphics[width=1.0\linewidth]{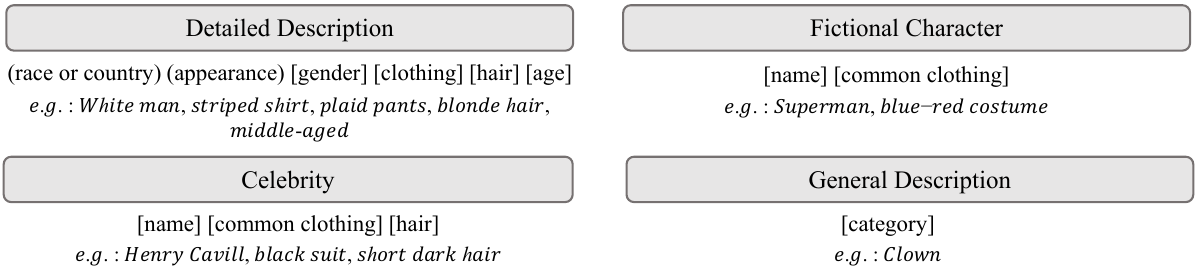}
   \caption{Prompt structure for ATLAS construction. }
   \label{image: gpt prompt}
   \vspace{-1.5em}
\end{figure}

\noindent\textbf{ChatGPT Prompt Structure.}
We divide avatar descriptions into four categories: detailed description, fictional character, celebrity, and general description. For each category, we design distinct generation templates. For each category in~\cref{image: gpt prompt}, ``[ ]'' content is included with every prompt, ``( )'' indicates the content appears randomly.

\noindent\textbf{Materials.} We provide more material settings for rendering textured humans. To achieve authentic human-like material, we set ``dielectric specular reflection'' to 0.1, and increase the ``Roughness'' to 0.6. Moreover, ``Sheen Tint'' is 0.5, ``Clearcoat Roughness'' is 0.03, and ``Index of refraction for transmission'' (IOR) is 1.45. The ``Alpha'' channel remains to be 1.

\begin{table}[b]
\centering
\vspace{-1.5em}
\caption{Overview of prompts we used for ATLAS sample texture image generation.}
\begin{tabular}{c|cc|cc} 
\cline{1-4}
                           & \multicolumn{2}{c|}{$T_{pos}$}                                                   & $T_{neg}$                                                                                                                                                                         &  \\ \cline{1-4}
\multirow{4}{*}{view} & \multicolumn{1}{l|}{front}     & \makecell{front side, from front, the \\ front of $T_{id}$}     & backside                                                                                                                                                                        &  \\ \cline{2-4}
                           & \multicolumn{1}{l|}{back}      & \makecell{backside, from back, the \\ backside of $T_{id}$}     & front, face, head                                                                                                                                                                 &  \\ \cline{2-4}
                           & \multicolumn{1}{l|}{left}      & left side, from left                       & front, back                                                                                                                                                                     &  \\ \cline{2-4}
                           & \multicolumn{1}{l|}{right}     & right side, from right                     & front, back                                                                                                                                                                     &  \\ \cline{1-4}
other                      & \multicolumn{2}{l|}{\makecell{black background, diffuse rendering, \\ daylight}}                                      & \makecell{overexposed, nude, layman work, \\ worst quality, teeth, smile, open mouth, \\ eyes closed} &  \\ \cline{1-4}
\end{tabular}
\captionsetup{skip=10pt}
\label{tab: prompts for datasets}
\end{table}

\section{More Analysis of TexDreamer}
\label{sec: More Analysis}

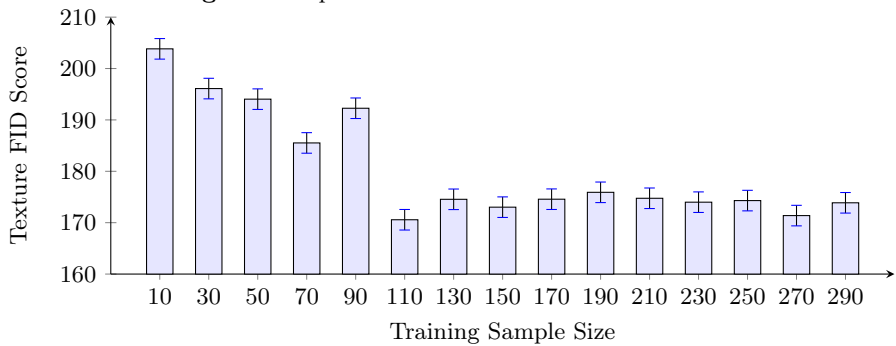
\begin{figure}[t]
\centering
\begin{tikzpicture}
\begin{axis}[
    ybar, 
    xlabel={Training Sample Size},
    ylabel={Texture FID Score},
    width=12cm,
    height=5cm,
    xmin=-10, xmax=310,
    ymin=160, ymax=210,
    xtick=data,
    axis lines=left, 
    ytick={160, 170, 180, 190, 200, 210},
    extra y tick style={grid=major, grid style={dashed, gray}}, 
    legend pos=north west,
]
\addplot+[fill=blue!10!white, draw=black, error bars/.cd, y dir=both, y explicit]
coordinates {
(10,203.83) +- (0,2)
(30,196.10) +- (0,2)
(50,194.04) +- (0,2)
(70,185.52) +- (0,2)
(90,192.27) +- (0,2)
(110, 170.57) +- (0,2)
(130,174.55) +- (0,2)
(150,173.02) +- (0,2)
(170,174.57) +- (0,2)
(190,175.91) +- (0,2)
(210, 174.75) +- (0,2)
(230, 173.99) +- (0,2)
(250, 174.30) +- (0,2)
(270, 171.38) +- (0,2)
(290, 173.87) +- (0,2)
};
\end{axis}
\end{tikzpicture}
\caption{Correlation between Training Sample Size and Texture FID Score.}
\label{fig: fid}
\vspace{-1.5em}
\end{figure}

\noindent\textbf{Training Sample Size.} Different training sample sizes can have different influences on model generation ability.  We use the FID score to evaluate different training sizes. FID score measures the similarity between the distribution of generated images and the distribution of real images. Lower FID values mean better image quality and diversity. Toward human texture generation from text, T2UV, we experiment with sample size spans from 10 to 300 with an interval of 20, see results in~\cref{fig: fid}. We find that the FID score between each test set and its train set tends to gradually start to become consistent over 100 training samples. This indicates that the model has reached a point of saturation with the given training data. In other words, the model learned the UV structure using around 100 textures, adding more training samples will not continue to contribute to improving model's ability to generate new, diverse, and high-quality textures.

\section{More Qualitative Results}
\label{sec:More Qualitative Results}

\begin{figure*}[t]
    \centering
    \includegraphics[width=1.0\textwidth]{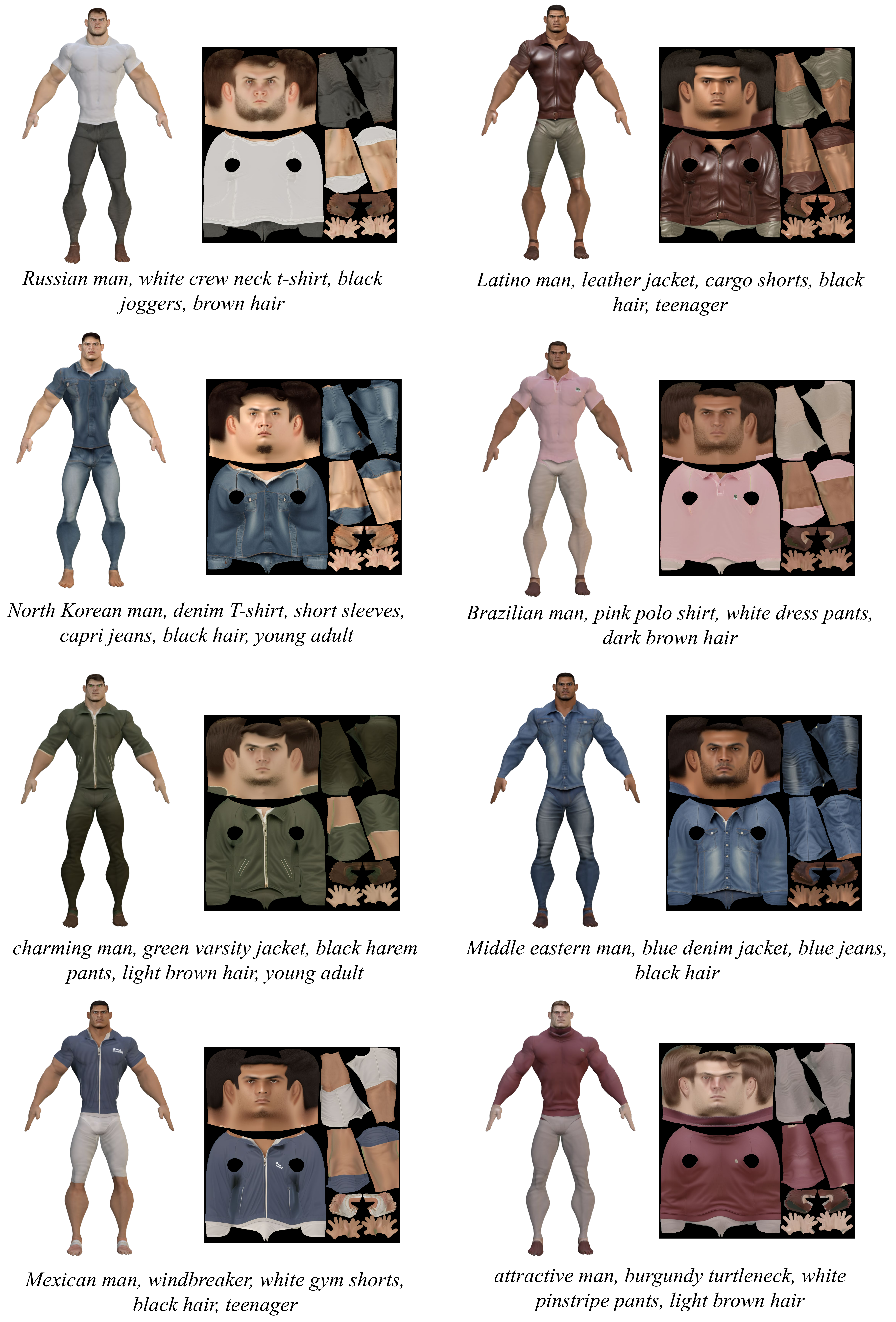}
    \caption{Realistic human textures generated from text with TexDreamer, each rendered with the same mesh. }
    \label{image: t2uv example 1 }
\end{figure*}

\begin{figure*}[htpb]
    \centering
    \includegraphics[width=1.0\textwidth]{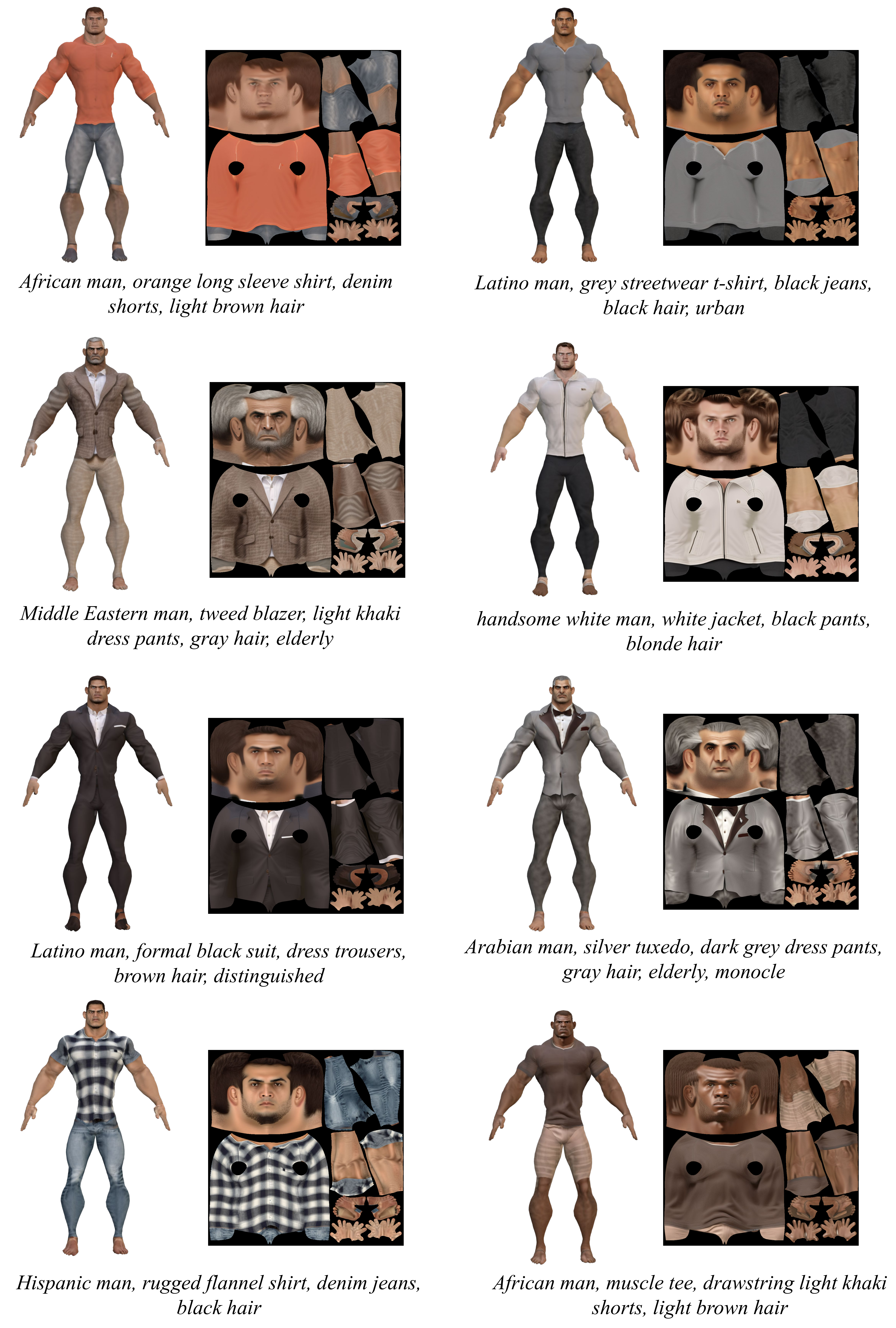}
    \caption{Realistic human textures generated from text with TexDreamer, each rendered with the same mesh. }
    \label{image: t2uv example 2 }
\end{figure*}

\begin{figure*}[htpb]
    \centering
    \includegraphics[width=1.0\textwidth]{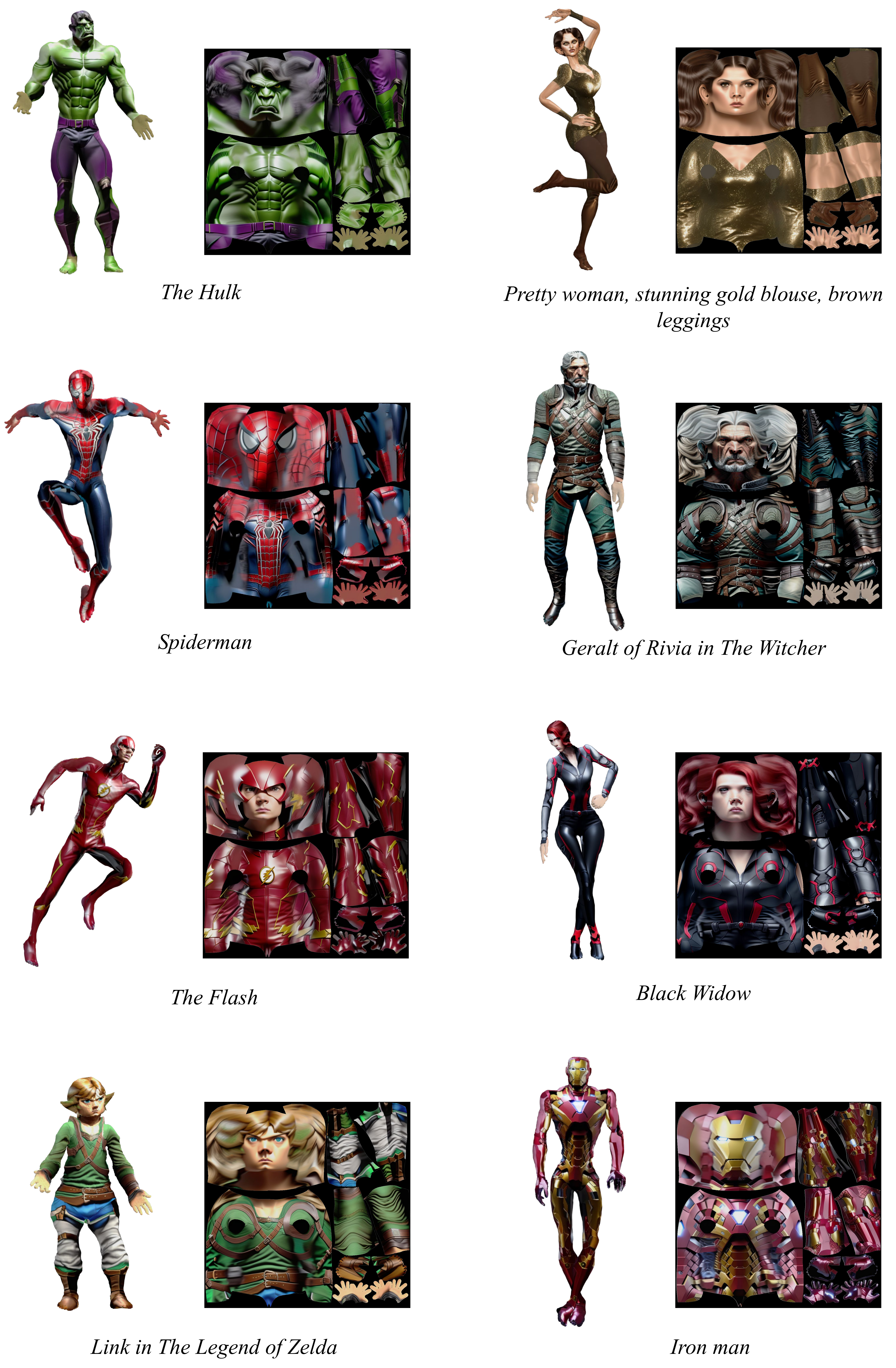}
    \caption{Fictional character textures generated from text with TexDreamer. }
    \label{image: t2uv example 3 }
\end{figure*}

\begin{figure*}[htpb]
    \centering
    \includegraphics[width=1.0\textwidth]{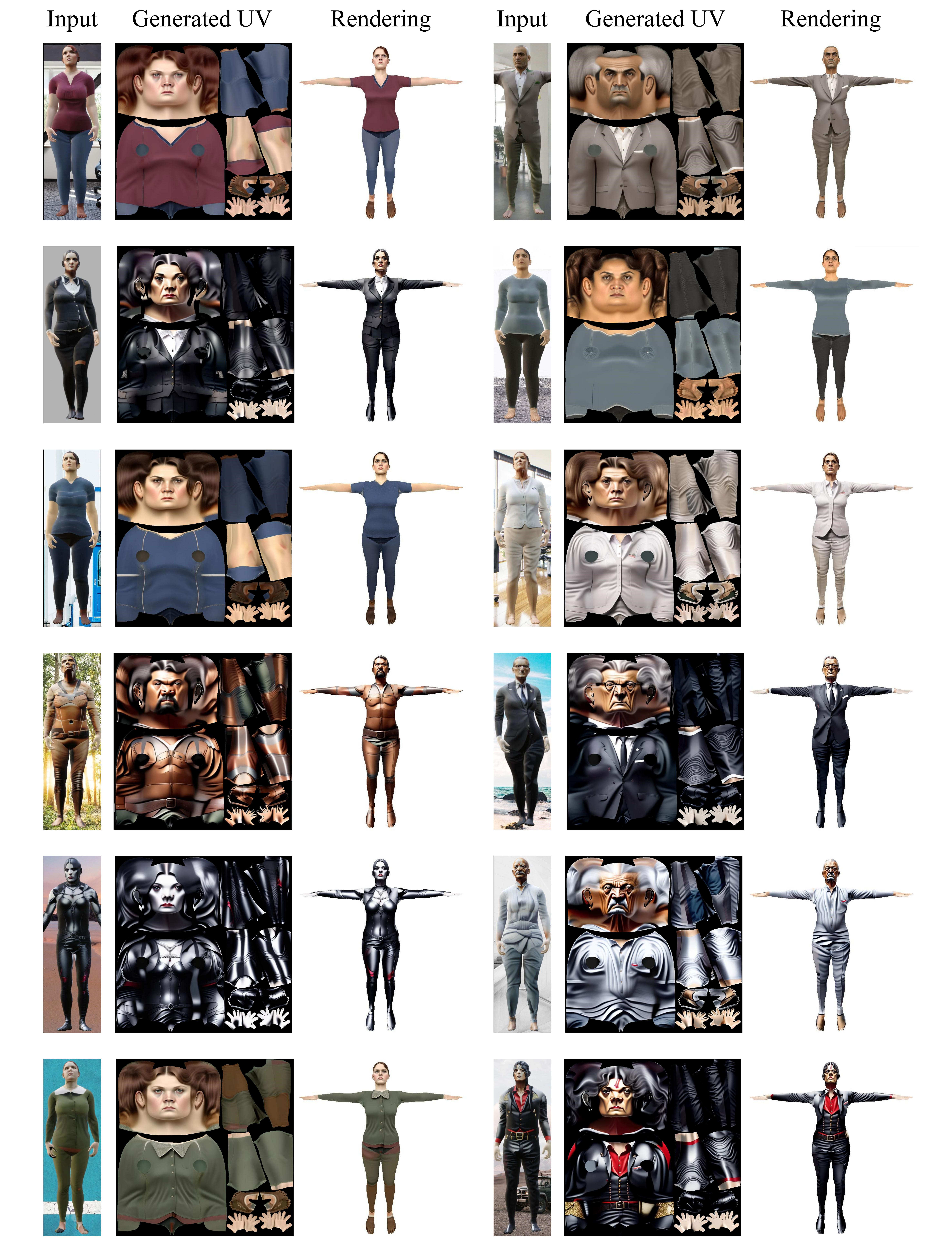}
    \caption{Texture generated from images with TexDreamer.}
    \label{image: i2uv example 1}
\end{figure*}

\begin{figure*}[htpb]
    \centering
    \includegraphics[width=1.0\textwidth]{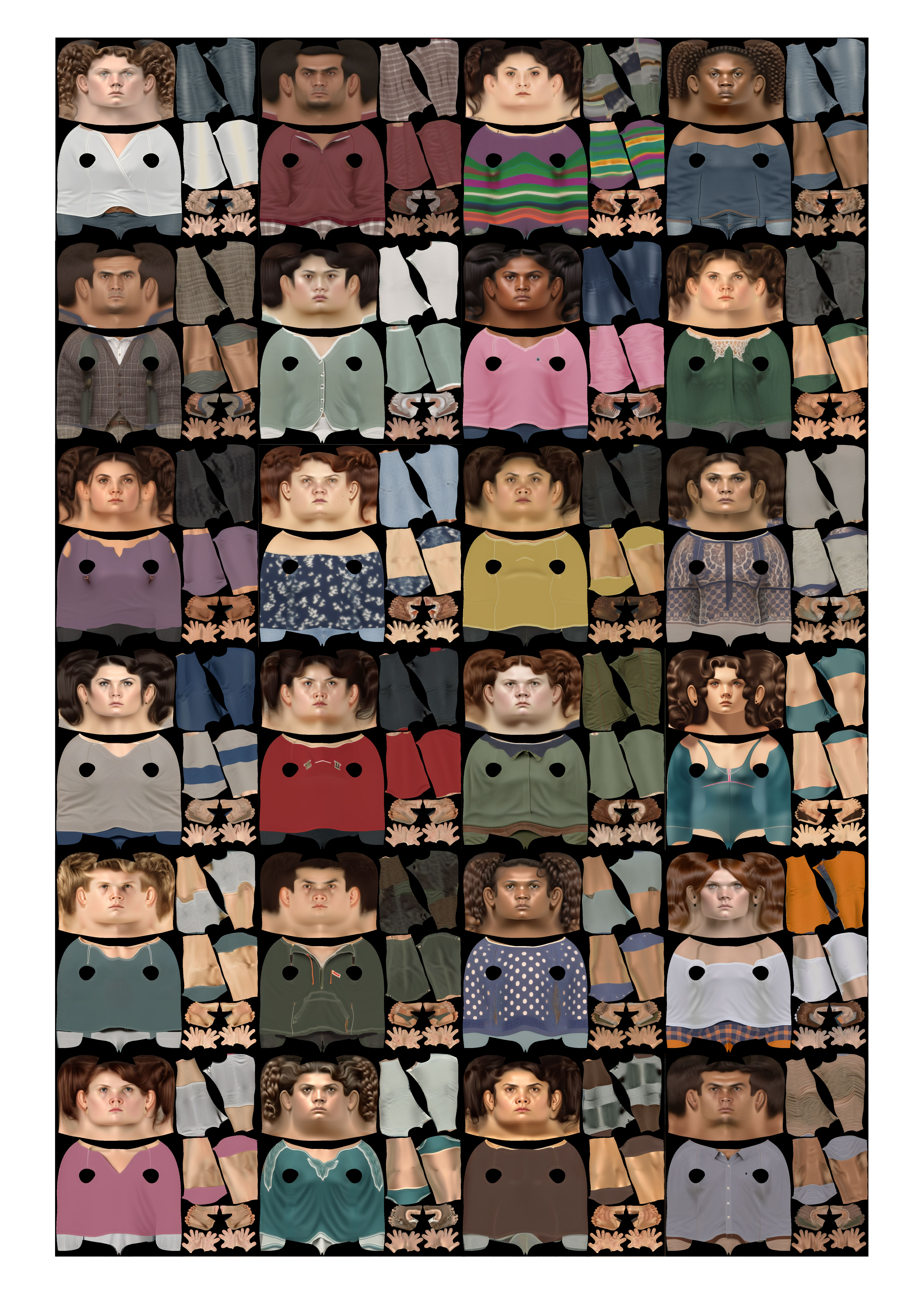}
    \caption{Human textures in our ALTAS dataset, example set 1. }
    \label{image: atlas example 1 }
\end{figure*}

\begin{figure*}[htpb]
    \centering
    \includegraphics[width=1.0\textwidth]{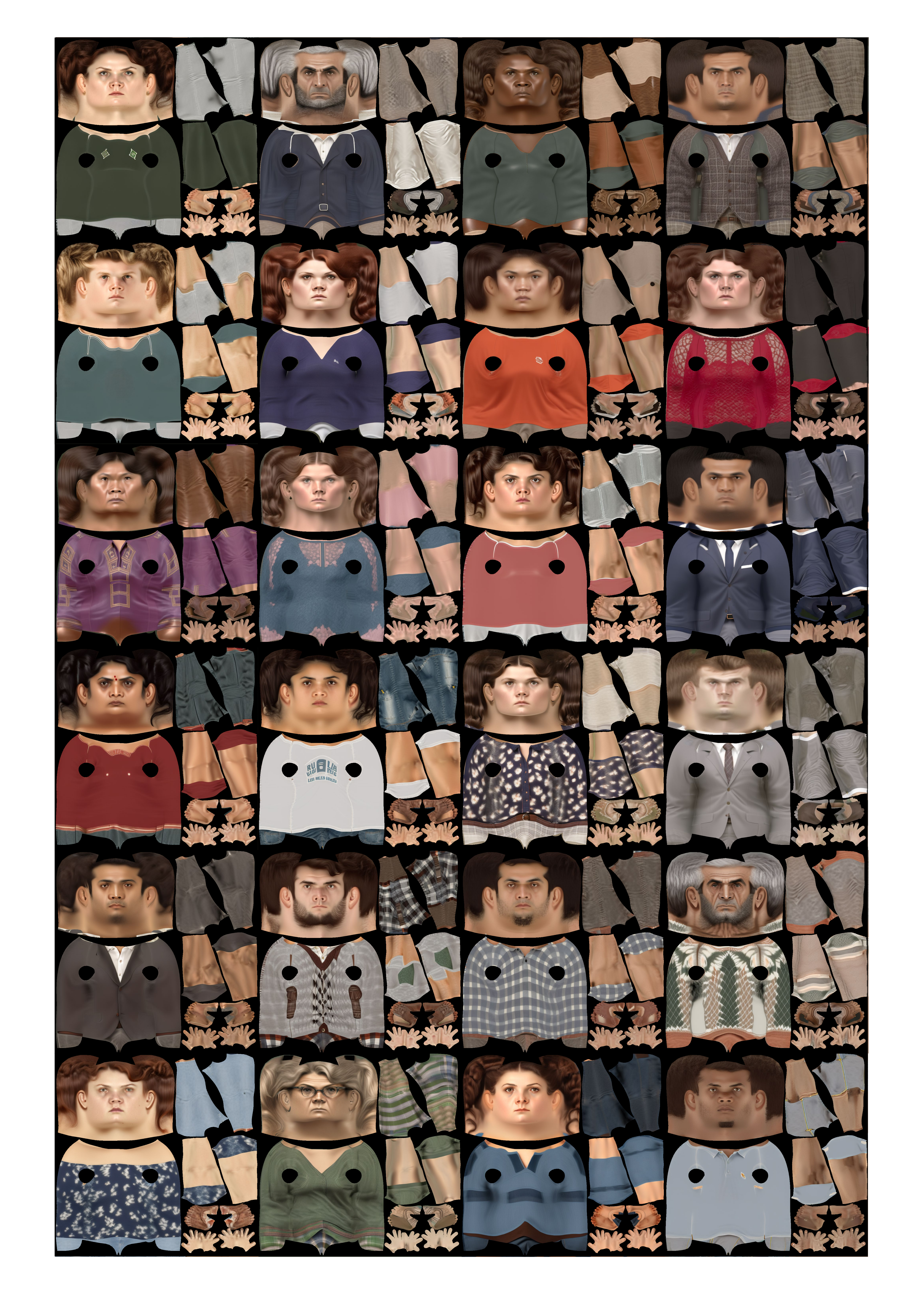}
    \caption{Human textures in our ALTAS dataset, example set 2.}
    \label{image: atlas example 2}
\end{figure*}

\begin{figure*}[htpb]
    \centering
    \includegraphics[width=1.0\textwidth]{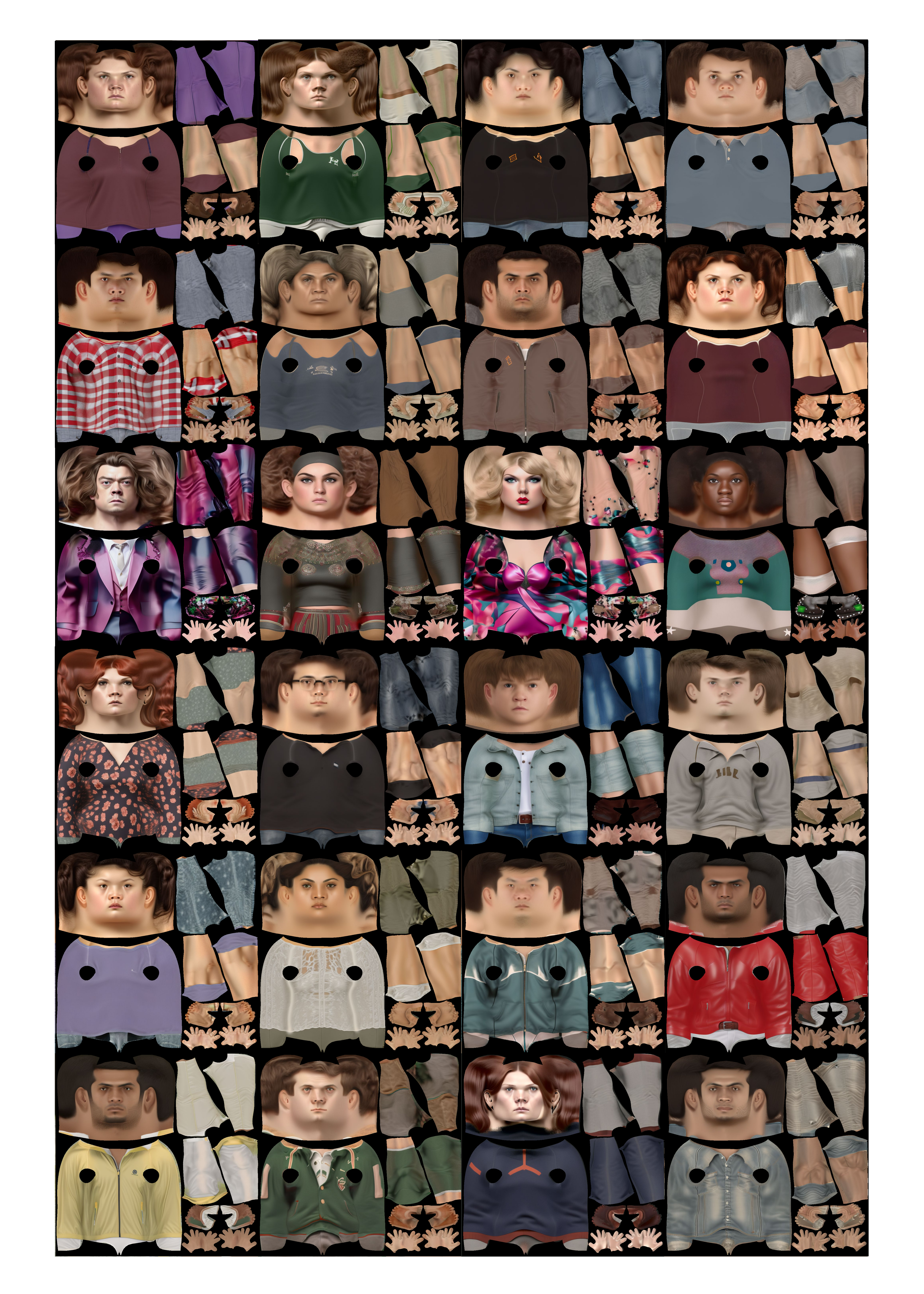}
    \caption{Human textures in our ALTAS dataset, example set 3.}
    \label{image: atlas example 3}
\end{figure*}

We show more qualitative results of TexDreamer and ATLAS. For generating textures from text, we show more results including both realistic humans and fictional characters in~\cref{image: t2uv example 1 },~\cref{image: t2uv example 2 }, and~\cref{image: t2uv example 3 }. Meshes are generated with text-to-avatar method TADA, and we animate the fictional characters with Mixamo.~\cref{image: i2uv example 1} also show more results with textures generated from images. Moreover, we display more human textures included in our ATLAS dataset, see~\cref{image: atlas example 1 },~\cref{image: atlas example 2} and~\cref{image: atlas example 3}.

\clearpage
\clearpage
\clearpage
\clearpage
\clearpage
\clearpage

\end{document}